\documentclass[10pt,letterpaper]{article}

\usepackage{cogsci}
\usepackage{pslatex}
\usepackage{apacite}
\usepackage{float}

\cogscifinalcopy % Uncomment this line for the final submission 

\usepackage{latexsym,amssymb,lastpage}
\usepackage{graphicx,amsfonts}
\usepackage{times,mathptmx,bm,amsmath}
\usepackage{xcolor}

\newenvironment{pf}[1][Proof]{\noindent\textit{#1. } }{\hfill$\square$}
  \newtheorem{thm}{Theorem}

\newcommand{\corr}[2]{{}{#2}}
\newcommand{\simocorr}[2]{{}{#2}}
\newcommand{\comment}[1]{{#1}}
\newcommand{\simo}[1]{{#1}}

\newcommand{\A}{\mathsf{A}}
\newcommand{\B}{\mathsf{B}}
\newcommand{\C}{\mathsf{C}}
\newcommand{\G}{\mathsf{G}}
\newcommand{\LL}{\mathsf{L}}
\newcommand{\M}{\mathsf{M}}
\newcommand{\X}{\mathsf{X}}
\newcommand{\Y}{\mathsf{Y}}
\newcommand{\Z}{\mathsf{Z}}

\title{Generalizing Outside the Training Set:\\ When Can Neural Networks Learn Identity Effects?}

\author{
 {\large \bf Simone Brugiapaglia (simone.brugiapaglia@concordia.ca)} \\
 {\large \bf Matthew Liu (matthew.liu@mail.concordia.ca)}\\
  Department of Mathematics, Concordia University, Montr\'eal, QC, Canada.\\
\AND
{\large \bf Paul Tupper (pft3@sfu.ca)} \\
Department of Mathematics, Simon Fraser University, Burnaby, BC, Canada.
   }

\begin{document}

\maketitle

\begin{abstract}
Often in language and other areas of cognition, whether two components of an object are identical or not determine whether it is well formed. We call such constraints \textit{identity effects}. When developing a system to learn well-formedness from examples,
it is easy enough to build in an identify effect. But can identity effects be learned from the data without explicit guidance? We provide a simple framework in which we can rigorously prove that \corr{a class of algorithms}{algorithms satisfying simple criteria} cannot make the correct inference. We then show that a broad class of algorithms including deep neural networks with standard architecture and training with backpropagation satisfy our criteria, \comment{dependent on the encoding of inputs}. Finally, we demonstrate our theory with computational experiments \corr{in which the choice of the encoding of inputs determines whether a given identity effect can be learned or not}{in which we explore the effect of different input encodings on the ability of algorithms to generalize to novel inputs}.

\end{abstract}

\section{Introduction}

Imagine subjects are told that the words $\A\A$, $\G\G$, $\LL\LL$, and $\M\M$ are good, and the words $\A\G$, $\LL\M$, $\G\LL$, and $\M\A$ are bad. If they are then asked whether $\Y\Y$ and $\Y\Z$ are good or bad, most will immediately say that $\Y\Y$ is good and $\Y\Z$ is bad. Humans will immediately note that the difference between the two sets of words is that the two letters are identical in the good words, and different in the second. The fact that $\Y$ and $\Z$ do not appear in the training data does not prevent them from making this judgement.

 However, many machine learning algorithms would not make this same inference given the training set. Depending on how inputs are provided to the algorithm and the training procedure used, the algorithm may conclude that since neither $\Y$ nor $\Z$ appears in the training data,  it is impossible to distinguish two inputs containing them.

The ability or inability of neural networks to generalize learning outside of the training set has been controversial for many years. \citeA{marcusbook} has made strong claims in support of the inability of neural networks and other algorithms that do not instantiate variables to truly learn identity effects and other algebraic rules. The explosion of interest in deep neural networks since that book has not truly changed the landscape of the disagreement; see \citeA{marcus2019} for a more recent discussion.
% \cite{marcusbook,berentbook}
Here we hope to shed some light on the controversy by considering a single instance of an algebraic rule, specifically an identity effect, and providing a rigorous framework in which the ability of an algorithm  to generalize it outside the training set can be studied. 

In our framework, we consider mappings that transform the set of inputs, and consider whether particular learning algorithms are invariant to these transformations, in a sense which we will define. We show that if both the learning algorithm and the training set are invariant to a transformation, then the predictor learned by the learning algorithm is also invariant to the transformation, meaning that it will assess inputs before and after transformation as equally well formed.
      We then show that a broad class of algorithms, including deep feedforward neural networks trained via backpropagation, satisfy our criteria \comment{for some commonly used encodings}. Finally, we show with computational experiments \corr{that a key issue for the applicability of the theorem is the encoding used}{how this dependence on encoding plays out in practice}. In our example we will see that one-hot encoding (also known as localist encoding) leads to a learner that is unable to generalize outside the training set, whereas distributed encoding allows partial generalization outside the training set. 

%The connectionist/symbolist dispute has been going on for a long time now.  Interestingly, the recent success of deep neural networks has not changed the landscape very much.

%Many of the arguments made for why connectionism cannot be a satisfactory model of human cognition can be found in the book \emph{The Algebraic Mind} \cite{marcusbook}. 

%These results tend to be of the qualitative variety. Though compelling they have not been formalized. Here we attempt to formalize them. 

%Our framework is based on the idea of \emph{a learning algorithm}. A learning algorithm takes a set of data we call the training set. For example, the training set may consist of a set of words each with a well-formedness rating.  

This work is a refinement and extension of earlier work \cite{tupper2016}. We have simplified the main theory and shown that it is applicable to a broader range of situations. \comment{Additionally, we have shown the theory applies to a large class of learning algorithms and encodings of inputs.}

\section{Main result}

%We consider a situation where we are 

Suppose we are training an algorithm to assign ratings to inputs. For example, we may want an algorithm that indicates whether a sentence is grammatical, whether two shoes in a picture form a matching pair, or decide whether a word is well-formed or not. Often the ratings will just be 0 or 1, like in the case of a binary classifier.
Let $W$ be the set of all possible inputs $w$. In our setting, $W$ is composed of words, but it may also consist of strings, vectors, images, etc.

Our learning algorithm is trained on a set of data $D$.
$D$ consists of a list of input-output pairs $(w,r)$ where $w \in W$ and $r \in \mathbb{R}$. Let $\mathcal{D}$ be the set of all possible data sets with words from $W$.

Typically, in machine learning there is a training algorithm (such as backpropagation, or least-squares fitting) which takes as input a training data set $D$ and outputs a set of parameters $p$. We formalize this with a map $\mathcal{A}$ as 
\[
p= \mathcal{A}(D).
\]
\simo{(Note that the training algorithm might involve randomized  operations, such as random parameter initialization; in this case, the set of parameters $p$ is a random variable).} Now, when we want to give a rating to a novel input $w$, we plug it into our model $f$ using the parameters $p$, i.e.
\[
r= f(p,w).
\] 
In the case of artificial neural networks, this operation  \simocorr{would correspond}{corresponds} to a forward propagation of $w$ through the trained network.

Though in practice determining $p$ is done separately from computing the rating of $w$ (especially since one usually wants multiple $w$ to be evaluated), for our purposes we can combine them into one function we can analyse. 
We define the  learning algorithm as a map $L \colon \mathcal{D} \times W \rightarrow \mathbb{R}$ given by
\[
L(D,w) = f(\mathcal{A}(D), w).
\]
%This is usually the form we expect $L$ to take, but it doesn't need to.

What we want to be able to show is that a given algorithm is not able to distinguish between two inputs not in $D$. More formally, we want our conclusion to be of the form
\[
L(D,w_1) = L(D,w_2),
\]
for two inputs $w_1, w_2$ in $W$, but not in $D$, when $L$ and $D$ have some particular structure.

The relation between $w_1$ and $w_2$ will be defined with the help of a function
 $\tau \colon W \rightarrow W$ that takes $w \in W$ and gives $\tau(w) \in W$. 
 It is some transformation of the inputs.
For example, if $W$ is a set of words, $\tau$ might reverse the order of the letters. If $W$ is a set of images, $\tau$ might perform a mirror reflection. In the case of a data set $D$, we define $\tau(D)$ as the data set obtained by replacing every instance of $(w,r)$ in $D$ with $(\tau(w),r)$.

Our main result follows.\ 

\begin{thm}[Rating impossibility for invariant learners]
Let $L$ be a learning algorithm, $D$ a data set, $w \in W$ an input, and $\tau$ a transformation of $W$. Assume that the following two conditions hold:
\begin{enumerate}
\item $L(\tau(D), \tau(w)) = L(D,w)$ (invariance of the algorithm); 
\item $\tau(D) = D$ (invariance of the data). 
\end{enumerate}
Then, $L(D,\tau(w)) = L(D,w)$.
\end{thm}

\begin{pf}
$$
L(D,\tau(w)) = L(\tau(D), \tau(w)) = L(D,w).
$$
\end{pf}

The first condition in the theorem, \textit{invariance of the algorithm}, 
%\comment{[Do we want to underline sentences? I personally find it a bit weird... I switched to $\backslash$textit$\{\}$. If you like, we can go back to $\backslash$emph$\{\}$.]}
 we will show to be true of some learning procedures for all $D$ and $w$, though the result only requires it for the $D$ and $w$ of interest. The second condition, \textit{invariance of the data}, we expect to hold only for certain particular data sets, and, in particular, the richer the data set, the fewer transformations $\tau$ it will be invariant to. Under these two conditions, the theorem states that the algorithm will not be able to give different ratings to $w$ and $\tau(w)$.

Here is a simple example of how this theorem works. Suppose $W$ consists of two-letter words and $\tau$ is a transformation that reverses the order of the two letters.
Suppose $L$ is a learning algorithm that is invariant to $\tau$, which is a fairly reasonable assumption, unless we explicitly build into our algorithm reason to treat either letter differently.
Suppose $D$ is a training set where all the words in it are just the same letter twice, so that $\tau(D) = D$. Then the theorem states that the learning algorithm trained on $D$ will give the same result for $w$ and $\tau(w)$ for all words $w$. So the algorithm will give the same rating to $xy$  and $yx$ for all letters $x$ and $y$. This is not surprising: if the algorithm has no information about words $xy$ where $x \neq y$, then why would it treat $xy$ and $yx$ differently?

Now we discuss how to apply this theorem to our actual motivating example, i.e.\ learning an identity effect.
%Our motivating application for this theory is an identity effect which is a little more involved.
%following, which addresses whether a learning algorithm can learn an identity effect and generalize it outside of its training space.
%The most interesting application of this result is currently:
 Again, suppose words in $W$ consist of ordered pairs of capital letters from the English alphabet.
 Suppose our training set $D$ consists of, as in our opening paragraph, a collection of two-letter words none of which contain the letters $\Y$ or $\Z$.  The ratings of the words in $D$ are $1$ if the two letters match and $0$ if they don't. 
%  Define the function $f \colon W \rightarrow \mathbb{R}$ by $f(xy)=1$ if $x=y$ and $f(xy)=0$ otherwise. So $f$ embodies the rule that checks whether two letters in a word are identical, and approves the word only if they are. We will use our theorem to show that a large class of algorithms are not able to do this task.
%
%Now suppose you have a learning algorithm and you train it on a $D$ containing all $(w,f(w))$ when $w$ does not include the letters Y or Z. Is the algorithm going to be able to determine the correct values for $\Y\Y$ and $\Y\Z$?
%Does the algorithm think that $f(\Y\Y)=f(\Y\Z)$?
To apply the theorem, let $\tau$ be defined by
\[
\tau(x\Y)=x\Z, \ \ \ \tau(x\Z)=x\Y, \ \ \ \tau(xy)=xy, 
\] 
for all letters $x$ and all letters $y$ with $y \neq \Y,\Z$. So $\tau$ usually does nothing to a word, but if the second letter is a $\Y$, it changes it to a $\Z$, and if the second letter is a $\Z$, it changes it to a $\Y$.
Note that since our training set  $D$ contains neither the letters $\Y$ nor $\Z$, then $\tau(D)=D$, as all the words in $D$ satisfy $\tau(w) =w$.

According to our theorem, to show that $L(D,\Y\Y)=L(D,\Y\Z)$, and therefore that the learning algorithm is not able to generalize the identity effect correctly outside the training set, we just need to show that
\[
L(\tau(D),\tau(w))= L(D,w),
\]
for our $D$ and $w=\Y\Y$. In fact we will show that this identity is true for all $D$ and $w$ for certain algorithms and encodings of the inputs.

\section{Encodings}

Up till now, we have let our set of inputs $W$ be any set of objects. But in practice, our inputs will always be encoded as vectors. We use $w$ to denote both the input and its encoded vector. We will also consider maps $\tau$ that are implemented by linear transformations when working with encoded vectors. We denote the linear transformation that implements $\tau$ by $\mathcal{T}$. 

As an example, the map $\tau$ we previously introduced, that switches $\Y$ and $\Z$ in the second position of a word, will be implemented by a linear transformation $\mathcal{T}$, but the particular transformation will depend on how we encode the two-letter words as vectors. We will obtain different results for the invariance of a learning algorithm depending on the properties of $\mathcal{T}$.

\section{Which learning algorithms are invariant?}

\subsection{No regularization}

We suppose our model for the data $D=\{(w_i,r_i)\}_{i = 1}^n$ is given by $r =f(B, C  w)$ where  $C$ is a matrix containing the coefficients multiplying $w$ and $B$ incorporates all other parameters including any constant term added to $C w$ (e.g., the first bias vector in the case artificial neural networks). The key point is that the parameters $C$ and the input $w$ only enter into the model through $C w$. 

This at first might seem restrictive, but in fact most neural network models use this structure: input vectors are multiplied by a matrix of parameters before being processed further. For example, suppose we are training a three-layer feedforward neural network whose output $r$ is given by
\[
r=\sigma_3( W_3 \, \sigma_2( W_2 \, \sigma_1( W_1 w +b_1) + b_2) +b_3),
\]
where $W_1, W_2, W_3$ are weight matrices, $b_1$, $b_2$, $b_3$ are bias vectors, and $\sigma_1, \sigma_2, \sigma_3$ are nonlinear activations (e.g., ReLU or sigmoid functions). 
In this case, we can let $C = W_1$ and $B=(W_2, W_3, b_1, b_2, b_3)$ to show that it fits into the required form.

Now suppose we select $B$ and $C$ by optimizing some loss function
\begin{equation}
\label{eq:loss_function}
F(B,C) = \mathcal{L}( f(B, C  w_i),r_i, i=1\ldots n),
\end{equation}
Let $\hat{B}$ and $\hat{C}$ be the optimal values of $B$ and $C$ and let us assume them, for the moment, to be unique minimizers.

Let us now assume that the transformation $\tau$ is linear and invertible, hence of the form $\tau(w) = \mathcal{T} w$, for some invertible matrix $\mathcal{T}$. If we apply $\mathcal{T}$ to the words $w_i$ in the data set and perform optimization again, we get new parameters $B'$ and $C'$.
But note that $C' (\mathcal{T} w_i) = (C' \mathcal{T}) w_i$. So the optimum is obtained by letting
\[
 C' \mathcal{T} = \hat{C},
\]
or $C' = \hat{C} \mathcal{T}^{-1}$, and $B' = \hat{B}$.

But what output do we get with these new parameters for the input $\tau(w)$? We obtain
\[
L(\tau(D), \tau(w)) = f(B',C' \mathcal{T} w)= f(B,\hat{C} w) = L(D,w),
\]
as required. The fact that $w$ is premultiplied by a matrix that is fit as part of the learning algorithm means that it doesn't matter whether all inputs are premultiplied by a linear transformation.

Summarizing these considerations we obtain the following theorem.
\begin{thm}
Suppose that a learning algorithm $L$ uses a model of the form $r = f(B,Cw)$, where parameters $B$ and $C$ are determined by minimizing a loss function of the form \eqref{eq:loss_function} and that admits a unique set of parameters $(\hat{B},\hat{C})$ as its global minimizer. Then, for any $D$ and $w$, $L$ is invariant to any $\tau$ that is a linear invertible transformation: 
$$
L(\tau(D), \tau(w))=L(D,w).
$$
\end{thm}

\subsection{Regularization}

So far we have considered a loss function where the parameters $C$ that we are fitting only enter through the model $f$ in the form $C w_i$. But, more generally, we may consider the sum of a loss function and a regularization term: 
%\comment{[For consistecy with the ML language, I called $\mathcal{L}$ the loss (and not $F$). I also replaced $K(B,C)$ with $\mathcal{R}(B,C)$ and included the $\lambda$ at this stage .]}
\[
F(B,C) = \mathcal{L}( f(B, C  w_i),r_i, i=1\ldots n)+ \lambda \mathcal{R}(B,C),
\]
where $\lambda > 0$ is a tuning parameter. Suppose $B$ and $C$ are obtained by minimizing this objective function.

Suppose our map $\tau$ is again implemented by an invertible matrix $\mathcal{T}$ in our encoding.
As long as $\mathcal{R}(B, C \mathcal{T}) =\mathcal{R}(B,C)$ for all $B$ and $C$ and if $F(B,C)$ still admits a unique set of minimizers, then the arguments of the previous subsection go through as before.
This begs the question: what linear transformations $\tau$ will make this true? If $\mathcal{R}$ has the form
\[
\mathcal{R}(B,C)= \mathcal{R}_1(B) + \| C\|^2_F,
\]
where $\| \cdot \|_F$ is the Frobenius norm (obtained by squaring and adding all the coefficients in $C$), also known as $L^2$ regularization, then any orthogonal transformation $\tau$ will lead to a learning algorithm that is invariant to $\tau$.

If we use an $L^1$ regularization term (obtained by summing the absolute value of all the entries in $C$), the algorithm will not be invariant to all orthogonal transformations, but it will be to $\tau$ that are implemented by a permutation matrix, as it is in our motivating example with localist encoding.

\subsection{Multiple minima and backpropagation}

It is an idealization of most learning algorithms to assume that they are trained by finding unique global minimizers of loss functions. Models are often not trained all the way to a minimum, there may be multiple minima, and there may be local, non-global minima.
In order to determine if a learning algorithm $L$ is invariant to a transformation $\tau$, we have to study how the parameters are actually learned from the data.

For deep neural networks, which are our focus here, a standard training method is backpropagation, which can be viewed simply as gradient descent. Parameters are determined by randomly generating initial guesses and then using gradient descent to find values that sufficiently minimize the loss function.

Let us consider a linear orthogonal transformation $\tau$ associated with a linear orthogonal matrix $\mathcal{T}$. We randomly initialize the parameters $C$ as $C = C_0$, such that $C_0$ and $C_0\mathcal{T}$ have the same distribution. This happens, for example, when the entries of $C_0$ are identically and independently distributed according to a normal distribution $\mathcal{N}(0,\sigma^2)$. (Note that this scenario includes the deterministic initialization $C_0 = 0$, corresponding to $\mathcal{N}(0,0)$). We also  initialize $B=B_0$ in some randomized or deterministic way independently of $C_0$.

The subsequent estimates of $C_{i+1}$ for $C$ are then computed via backpropagation as
$$
C_{i+1} = C_{i} + \theta_{i} \frac{\partial F}{\partial C} (B_i, C_i),
$$
for $i = 0,1,\ldots,k$ and a sequence of step sizes $\{\theta_i\}_{i = 1}^k$, which we assume to be independent of $C_0$. Successive approximations $B_i$ of $B$ are computed similarly. 

Now, what happens if we apply the \textit{same} training strategy using the transformed data set $\tau(D)$? We denote the generated parameter sequence with this training data $\{(B'_i,C'_i)\}_{i = 1}^k$.
We claim that the sequence $(B'_i,C'_i \mathcal{T})$ has the same distribution as $(B_i,C_i)$ for all $i$. Then, if we use $(B_k,C_k)$ as the parameters in our model we obtain
$$
L(\tau(D),\tau(w)) = f(B'_k,C'_k \mathcal{T} w ) 
$$
which has the same distribution as $L(B_k,C_k w)$, establishing invariance of the learning algorithm to $\tau$.

The full statement of our results is as follows; we provide a full proof in another publication.

\begin{thm}
Let $\tau$ be a linear transformation with orthogonal matrix $\mathcal{T}$.
Suppose a learning algorithm $L$ uses a model of the form $f(B,Cw)$ and parameters $B$ and $C$ are determined by performing a predetermined number of gradient descent iterations to minimize an objective function of the form
\[
F(B,C)= \mathcal{L}(f(B,Cw_i),r_i, i=1,\ldots,n) + \lambda(\mathcal{R}_1(B) +  \|C\|^2_F).
\]
 Suppose the random initialization of the parameters $B$ and $C$ are independent and that the initial distribution of $C$ is invariant with respect to right-multiplication by $\mathcal{T}$.
Then, $L(D,w)$ and $L(\tau(D),\tau(w))$ have the same distribution. 
\end{thm}

\section{Numerical experiments}

Since our theoretical results apply to idealizations of the commonly used learning algorithms, here we explore how applicable they are with some numerical experiments. Our experimental setting is analogous to the one in \cite{tupper2016}. However, we will consider  different training algorithms and letter encodings.

\subsection{Task and data set}

Our vocabulary $W$ is the set of all two-letter words composed by any possible letter from $\A$ to $\Z$.  We define the set $W_1$ as the set of all grammatically correct words (i.e. $\A\A$, $\B\B$, ..., $\Z\Z$) and $W_0$ as the set of all other possible words (which in turn are grammatically incorrect). 

The training data set consists of the 24 words $\A\A$, $\B\B$, $\C\C$, ..., $\X\X$ from $W_1$ along with 48 words uniformly sampled from $W_0$ without replacement. The learners are then validated on the words $\Y\Y, \Z\Z, \Y\Z, \Z\Y, x\Y, x\Z$,
where $x \in \{\A,\B,...,\X\}$. We assign ratings 1 to words in $W_1$ and 0 to words in $W_0$.

%\comment{[Any reason twice as many bad words as good words?]} \comment{[S: We just considered the same experimental setting as in \cite{tupper2016}.]}

\subsection{Encodings}

We represent each word as the concatenation of the encodings of its two letters, and so the representation of the words is determined by the representation of the letters.
All letter representations used have a fixed length of $k = 26$ (chosen due to the 26 letters that make up our vocabulary $W$). We define one deterministic encoding and two random ones.

\textit{One-hot encoding} (or \textit{localist encoding}) is our sole deterministic encoding. This encoding simply assigns a single nonzero bit for each character. Namely, the letters $\A$ to $\Z$ are encoded using the standard basis vectors $\{e_i: 1 \leq i \leq k\}$ where $e_i$ has a 1 in position $i$ and 0's elsewhere. Because of its deterministic nature, new encodings are not generated at each repetition. 

\textit{Binary (or distributed) encoding} defines an arbitrary combination of $k$ bits as our representation, with all characters encoded uniquely. We also define a $j$-active bits binary encoding where only $j$ arbitrary bits are 1's with all  0. For our experiments, we set $j = 3$. Again, all characters are ensured to be encoded uniquely. Both one-hot and binary encodings are binary representations as all $k$ entries are constrained to be 1 or 0. 

% In the normal encoding, each character representation is sampled from the multivariate Gaussian distribution $X \sim \mathcal{N}(\mu,\,\tau)$. In our case, $\mu = 0$ and $\tau$ is the $k \times k$ identity matrix $I$. As such, each letter is represented as a random vector $x = (x_1, ..., x_k)$ where each entry $x_i$ is sampled from the $\mathcal{N}(0,\,1)$ distribution. \\

Finally, the \textit{Haar encoding} uses the rows of a random $k \times k$ matrix sampled from the orthogonal group $O(k)$ via the Haar distribution  \cite{mezzadri2006generate} for the representation of each of the $k$ letters. In other words, the row vector $a_i$ of the randomly sampled matrix $A$ is used for the representation of the $i$th letter, where $1 \leq i \leq 26$. Use of the Haar distribution ensures all encoded vectors are orthogonal to each other. 

In the context of our experiments, all random encodings are randomly re-generated for each repetition, producing new representations for each iteration.

Note that with these different encodings the map $\tau$ has representations as the matrix $\mathcal{T}$ with different properties. With the one-hot encoding, $\mathcal{T}$ is a permutation matrix (and hence orthogonal) that just switches the last two entries of a vector. With the Haar encoding, $\mathcal{T}$ is an orthogonal matrix. Finally, with the 3-active bit binary, $\mathcal{T}$ does not have any special algebraic properties.

\subsection{Neural network learners}

The learners we test are artificial feedforward neural networks with 1, 2 and 3 hidden layers.  Each hidden layer contains 256 units, with ReLU nonlinearities for all hidden units and a sigmoid activation for the output unit. All weights are initialized using the random Gaussian distribution $\mathcal{N}(\mu,\,\sigma^2)$ with $\mu = 0$ and $\sigma^2 = 0.0025$. Biases are initialized to 0. 

We train the models by minimizing the binary cross-entropy loss function via backpropagation using the Adam optimizer \cite{kingma2014adam} with the following hyperparameters: $\gamma = 0.001$, $\beta_1 = 0.9$ and $\beta_2 = 0.999$. The batch-size is set to 72 (the number of training samples) to ensure deterministic iterates and the number of epochs are tested at 100 and 500. The neural network architectures are implemented in Keras \cite{chollet2015keras}.

\begin{figure*}[ht]
\begin{center}
    \includegraphics[width=\textwidth]{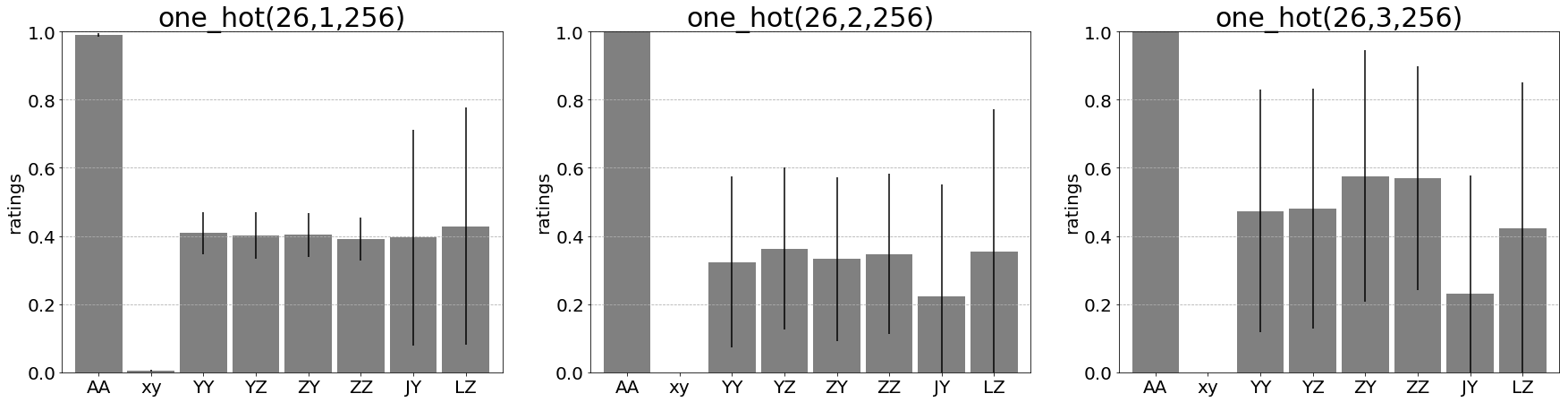}  
    \includegraphics[width=\textwidth]{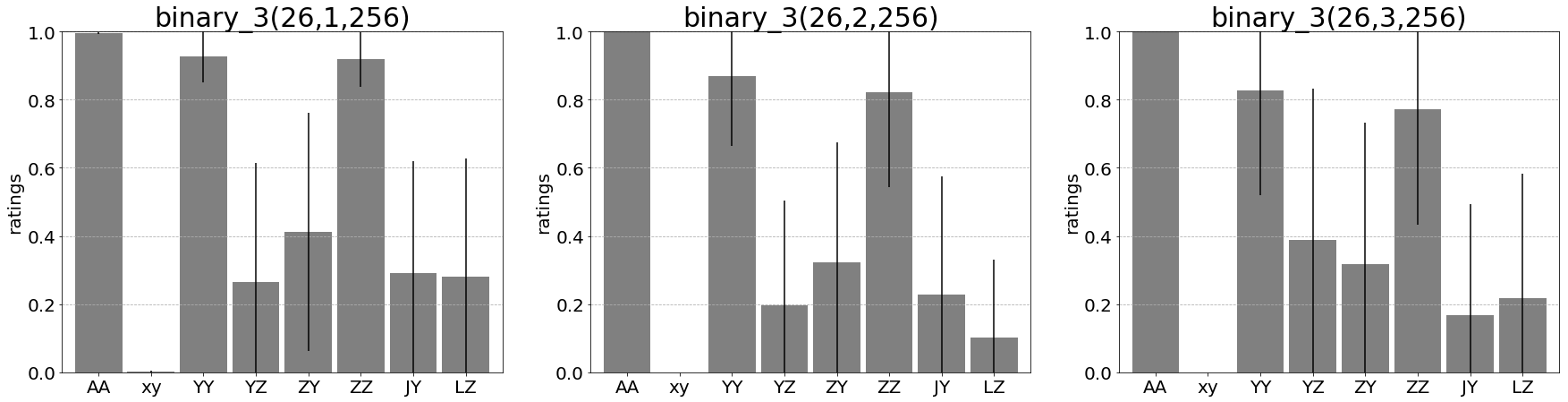}
    \includegraphics[width=\textwidth]{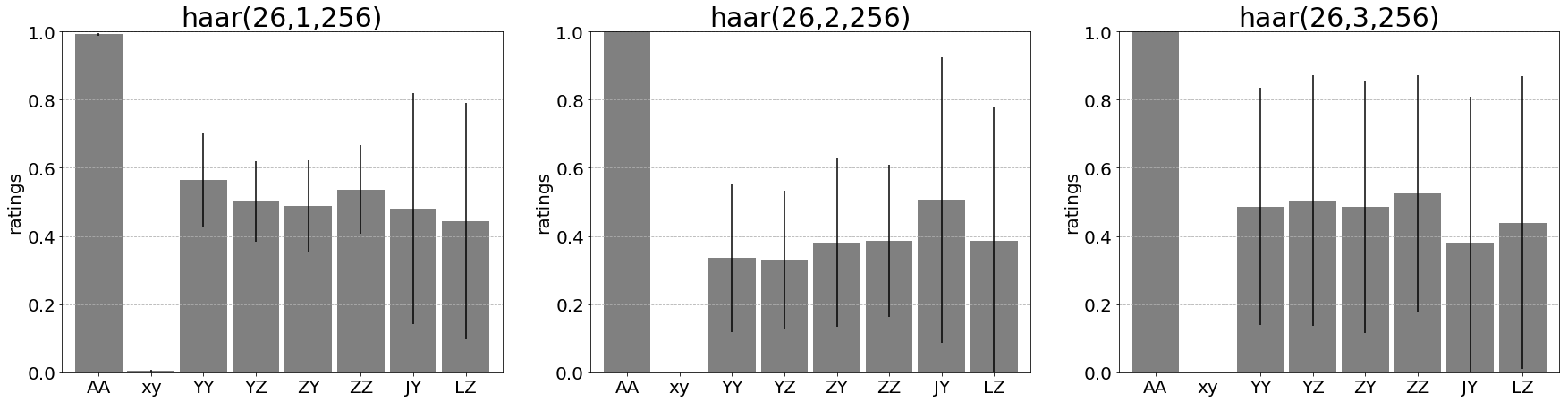}
    \caption{\label{fig:boxplots}Predictions of network architectures and encodings on novel words. From top to bottom: One-hot encoding, distributed 3-active bit encoding, and Haar encoding. From left to right: 1, 2, and 3 hidden layers.}
    \end{center}
\end{figure*}

\subsection{Randomness}

The experiment is repeated 40 times for each learner. For each iteration of the experiment, we randomly generate a new training data set. The validation data set is instead generated only once and held constant across all learners and all experiments.  For each encoding, the three neural network architectures are trained and validated in succession. Similarly, the sequence of 40 initial weights are also the same among all encodings. To further ensure consistency, the same random seed is set once at the beginning of each learner's experiment (not during the 40 individual experiments). As such, the runs for each encoding use the same sequence of 40 training data sets (and repeated 3 times for each architecture).

\subsection{Results}
We present the performance on a test set of each neural network architecture on each encoding in Figure~\ref{fig:boxplots}. 

The outputs shown correspond to a training session of 500 epochs. The first 2 bars of each graph correspond to words included in the training set ($xy$ denotes the first word from $W_0$ in the training set of a particular run). The boxes represent the average rating over all 40 outputs and the bars represent the corresponding standard deviation.

\begin{figure*}[ht]
\begin{center}
    \includegraphics[width=5.8cm, height = 4.3cm]{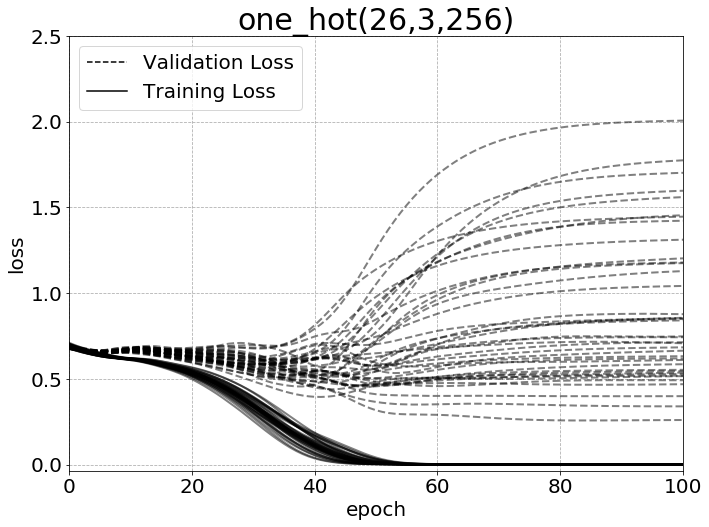}
        \includegraphics[width=5.8cm, height = 4.3cm]{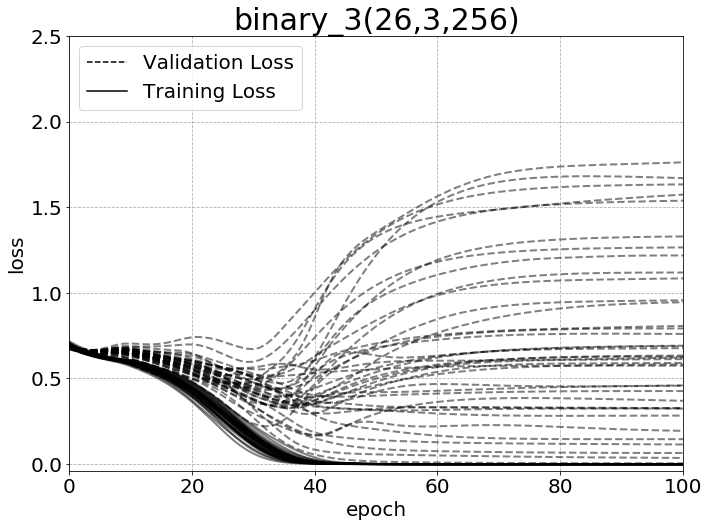}
    \includegraphics[width=5.8cm, height = 4.3cm]{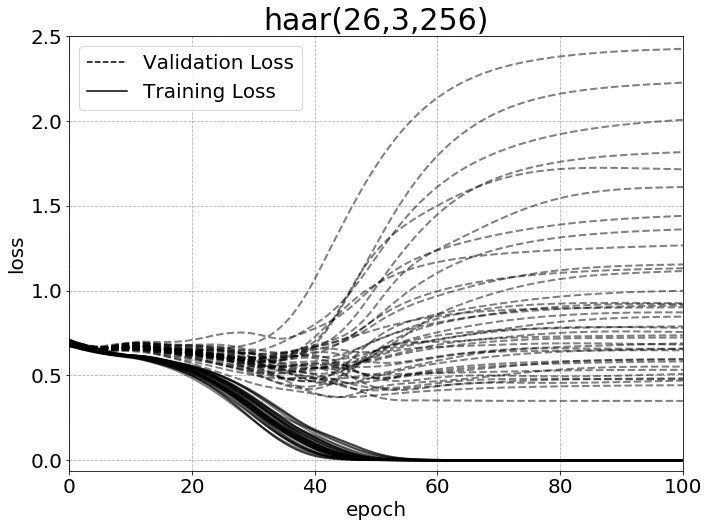}
    \caption{\label{fig:losses}Evolution of training loss (solid lines) and validation loss (dashed lines) of networks trained on the different encodings (3-layer case). From left to right: One-hot encoding, distributed 3-active bit encoding, and Haar encoding.}
    \end{center}
\end{figure*}

The box plots show that the neural networks are able to generalize (albeit not perfectly) to novel inputs on 1 of the 3 encodings tested, namely the binary 3-active bit encoding. This conclusion stems from the fact that higher than average scores are given by those learners to the novel stimuli $\Y\Y$ and $\Z\Z$ than to the novel stimuli $\Y\Z$, $\Z\Y$. The networks trained using the one-hot and Haar encodings show no discernible pattern indicating a complete inability to generalize the identify effects outside the training set. These results follow after all networks are observed to learn the training examples all but perfectly (as evidenced by the high ratings for column $\A\A$ and low ratings for column $xy$). 

It is interesting to note that both the one-hot and Haar encodings represent the only true orthogonal encodings. In the one-hot case, failure to generalize can be explained by the fact that the novel inputs stimulated connections and units that were never activated during the training phase. The Haar case addresses this issue by assigning nonzero values for each entry in the representation. However, it too fails to allow the networks to learn the identity effects. 

Figure~\ref{fig:losses} shows the evolution of the training and validation losses for the first 100 epochs when the 3-layer network is trained on different encodings. A discernible gap between dotted and solid lines indicate inability of the network to generalize to new inputs. The gap being present in one-hot and Haar cases while not being present in the distributed case mirrors the results we observe in Figure 1.

\section{Discussion} 

We see agreement between our theoretical predications and our numerical experiments for our identity effect test problem. Our theory predicted that when the encoded letters for different vectors are orthogonal (as they are with one-hot and Haar encodings), then since the transformation $\tau$  is an orthogonal transformation, the learner will not be able to distinguish between the inputs $\Y\Y$ and $\Y\Z$. The theory has nothing to say about the case of the 3-bit active encoding, because in that case $\tau$ is not orthogonal, and our theorems do not apply. Accordingly, in this case, even though the network is not able to give the correct answer of $1$ for $\Y\Y$ and $0$ for $\Y\Z$, and so not be said to learn the generalization perfectly, it does give a higher rating on average to  $\Y\Y$ than to $\Y\Z$.
We leave it to the reader to decide if this constitutes an exception to the claim that learners need to instantiate variables in order to generalize algebraic rules outside the training set \cite{marcus1999}.

Our results hew closely to those of \citeA{prickett2019learning}; see also \cite{prickett2018seq2seq}. There the authors train a variable-free neural network to perform reduplication, the process where a linguistic element is repeated from the input to the output. Following the experimental work of \citeA{marcus1999}, they trained the network on many examples of the pattern ABB, where A and B are substituted with syllables. The network is then tested by seeing if it can predict that the third syllable of a string such as ``li na $\rule{0.35cm}{0.15mm}$'' should be ``na'', even when not exposed to this input before. The authors found that their network could perform partial generalization when the novel inputs included new syllables or new segments, but could not generalize to new feature values. The reason for this is that feature values were encoded in their model via a localist representation, and introducing a new feature value was like expecting the network to learn a function depending on a bit that was always set to zero in the training data, just like the localist representation in our set-up. Since novel segments were composed of multiple novel feature values, this corresponds to our 3-bit active encoding, where apparently learning can be extended imperfectly to new combinations of already seen segments.

\comment{Our results and those of \citeA{prickett2019learning} continue a theme that is well known in connectionist literature: when representations on novel inputs overlap with representations in training data, networks are able to generalize training to novel inputs. See \citeA{mcclelland1999} for a discussion of this point in the context of identity effects.}

\section{Acknowledgments}

The authors were supported by NSERC Discovery Grants. \simo{S.B.\ and M.L.\ also acknowledge the Faculty of Arts and Science of Concordia University for financial support.}

\bibliographystyle{apacite}

\setlength{\bibleftmargin}{.125in}
\setlength{\bibindent}{-\bibleftmargin}

\end{document}